\relax
\documentclass{llncs}

\usepackage{times}
\usepackage[scaled=.92]{helvet}
\usepackage{courier}

\usepackage{url}  \usepackage{booktabs}       \usepackage{amsfonts}       \usepackage{nicefrac}       \usepackage{microtype}      

\usepackage{amsmath}
\usepackage{amssymb}

\usepackage{graphicx}

\usepackage{bbold}
\usepackage{ntheorem}

\usepackage{subfig}
\usepackage{multirow}

\usepackage[x11names]{xcolor}

\newtheorem{Proposition}{Proposition}

\DeclareMathOperator{\E}{E}     \DeclareMathOperator{\Var}{Var} \DeclareMathOperator{\diag}{diag}
\def\mat#1{\mathbf{#1}}

\let\methodfont\textrm
\def\movequery{\methodfont{Move-query}}
\def\movelabeled{\methodfont{Move-labeled}}

\title{A Fast and Easy Regression Technique for $k$-NN Classification Without Using Negative Pairs\thanks{Earlier version of this paper appeared in PAKDD '17.  This version corrects an error in Eq.~\eqref{eq:closed-form}.}}

\author{
  Yutaro Shigeto
  \and Masashi Shimbo
  \and Yuji Matsumoto
}
\institute{
Nara Institute of Science and Technology, Ikoma, Nara, Japan
\email{\{yutaro-s,shimbo,matsu\}@is.naist.jp}
}

\begin{document}

\maketitle

\begin{abstract}
  This paper proposes an inexpensive way to learn an effective dissimilarity function to be used for
  $k$-nearest neighbor ($k$-NN) classification.
  Unlike Mahalanobis metric learning methods that map both query (unlabeled) objects
  and labeled objects to new coordinates
  by a single transformation,
  our method learns a transformation of labeled objects to new points in the feature space
  whereas query objects are kept in their original coordinates.
  This method has several advantages over existing distance metric learning methods:
  (i) In experiments with large document and image datasets,
  it achieves $k$-NN classification accuracy better than or at least comparable
  to the state-of-the-art metric learning methods.
  (ii) The transformation can be learned efficiently by solving a standard ridge regression problem.
  For document and image datasets,
  training is often more than two orders of magnitude faster than the fastest metric learning methods tested.
  This speed-up is also due to the fact that the proposed method eliminates the optimization over
  ``negative'' object pairs,
  i.e., objects whose class labels are different.
  (iii) The formulation has a theoretical justification in terms of reducing hubness in data.
 \end{abstract}

\section{Introduction}
\label{sec:introduction}

Let $\mathcal{X}$ be a feature space and $\mathcal{Y}$ be a set of class labels.
The \emph{$k$-nearest neighbor} ($k$-NN) classifier
predicts the class label of an unknown query object $\mat{x} \in \mathcal{X}$ by its nearest neighbors.
Given $\mat{x}$ and
a set of labeled objects $\mathcal{D} = \{ (\mathbf{x}_i, y_i) \}_{i=1}^{n}$
where $\mat{x}_i \in \mathcal{X}$ is the feature vector of the $i$th object
and $y_i \in \mathcal{Y}$ its class label,
the classifier first computes the distance between $\mathbf{x}$
and each labeled object $\mathbf{x}_i$.
It then predicts the class label $\hat{y}$ of $\mathbf{x}$ by the majority among its $k$ nearest labeled objects.
When $k=1$, the decision rule of the $k$-NN ($1$-NN) classifier is simply:
\begin{equation}
  \hat{y} = \mathop{\arg \min}_{y_i : (\mathbf{x}_i, y_i) \in \mathcal{D}} f (\mathbf{x}, \mathbf{x}_i) , \label{eq:1-nn}
\end{equation}
where function $f : \mathcal{X} \times \mathcal{X} \rightarrow \mathbb{R}$ is some distance/dissimilarity function.

Obviously, the choice of function $f$ affects the accuracy of classification. 
Therefore,
many researchers \cite{Weinberger2009,Davis2007,Ying2012,Xing2002} have tackled \emph{metric learning},
which is the task of learning a suitable distance function
from data.

For Euclidean object space $\mathcal{X}=\mathbb{R}^d$,
metric learning is usually formulated as the task of finding
a Mahalanobis distance.
In this formulation,
the distance between two objects $\mat{x}, \mat{z} \in \mathbb{R}^d$ is defined by
\begin{equation}
  f(\mat{x}, \mat{z})
  = \sqrt{ (\mat{x}-\mat{z})^{\text{T}} \mat{M} (\mat{x} - \mat{z}) }, 
  \label{eq:mahalanobis-distance-orig}
\end{equation}
with some positive (semi)definite matrix $\mat{M}$.
By defining matrix $\mat{L}$ by $\mat{M}=\mat{L}^{\text{T}} \mat{L}$,
we can write
the distance in Eq.~\eqref{eq:mahalanobis-distance-orig}
as
\begin{equation}
  f(\mat{x}, \mat{z})
  = \| \mat{L}\mat{x} - \mat{L}\mat{z}\|.
  \label{eq:mahalanobis-distance}
\end{equation}
This equation shows that learning Mahalanobis distance is equivalent to learning a suitable linear transformation $\mat{L}$.

In the context of $k$-NN classification,
distance needs to be measured only between query (unlabeled) objects and labeled objects,
as can be seen from Eq.~\eqref{eq:1-nn};
when distance $f(\mat{x}, \mat{z})$ is computed, the first object $\mat{x}$ is always a query object,
and the second object $\mat{z}$ is always a labeled object $\mat{x}_i$.
Moreover, function $f$ need not be metric and can be any measure of dissimilarity;
for instance, $f$ being asymmetric is perfectly acceptable.

In this paper, 
we learn one such dissimilarity function.
The idea is to compute a transformation of labeled objects to new points
while unlabeled objects are kept at their original points.
Thus, our objective is to find a suitable matrix $\mat{W}$ that defines a dissimilarity
function
\begin{align}
  f(\mat{x}, \mat{z}) &= \| \mat{x} - \mat{W}\mat{z}\|,
                        \label{eq:backward-mapping}
\end{align}
where $\mat{x}$ is a query object, and $\mat{z}$ is a labeled object.

Because the coordinates of query objects are fixed,
our formulation might appear less flexible than Mahalanobis distance learning
(Eq.~\eqref{eq:mahalanobis-distance}).
However, as shown in a subsequent section,
it gives a better $k$-NN classification accuracy than
Mahalanobis distance learning methods on many datasets
that feature high-dimensional space.
Moreover,
optimizing $\mat{W}$ in Eq.~\eqref{eq:backward-mapping} is much easier and substantially (often more than two orders of magnitude) faster.

The effectiveness of the proposed approach has a theoretical foundation
in terms of reducing \emph{hubness} in data \cite{Radovanovic2010,Shigeto2015}.
Recent studies have shown that the presence of hubs, which are a few objects that appear in the $k$-NNs of many objects,
is an obstacle that can
harm the performance of many vector space methods
\cite{Radovanovic2010,Schnitzer2012,Suzuki2013}. We show that metric learning is no exception,
and the transformation of labeled objects restrains hubs from emerging.
This approach is justified by a recent result
of Shigeto et~al. \cite{Shigeto2015},
who used regression to reduce hubness in zero-shot problems.
In their work, the problem was cast as a task of cross-domain matching,
whereas in this paper, we are concerned with improving the accuracy of
$k$-NN classification in a single space.

Another notable feature of the proposed method is that it
eliminates optimization over ``negative'' object pairs, i.e., objects belonging to difference classes.
In other words, our method
only attempts to make
objects of the same class (``positive'' object pairs) to be closer.
Its objective function does not have any constraints or terms that promote
negative object pairs to be apart from each other.
Such constraints are indispensable in Mahalanobis metric learning
to prevent trivial solutions $\mat{M} = \mat{O}$
in Eq.~\eqref{eq:mahalanobis-distance-orig} or $\mat{L} = \mat{O}$ in Eq.~\eqref{eq:mahalanobis-distance},
and metric learning typically optimizes over a large number of negative object pairs.
Moreover, incorporating negative pairs results in a non-convex optimization problem
with respect to matrix $\mat{L}$.
Existing metric learning methods
\cite{Weinberger2009,Ying2012,Davis2007,Jain2012,Xing2002}
hence resorts to optimizing $\mat{M}=\mat{L}^{\text{T}}\mat{L}$
using computationally intensive methods such as semi-definite programming.
By contrast,
since we only transforms labeled objects, we need not worry about $\mat{W}=\mat{O}$ being the solution
(see Eq.~\eqref{eq:backward-mapping}),
thus eliminating the need of negative pairs.
This makes the solution easily obtained with ridge regression,
which contributes to reduced computation time.

\section{Related work}
\label{sec:related-work}

We briefly review some of the metric learning methods,
mostly those used in the experiments in Section~\ref{sec:experiments}.
For comprehensive survey of the field, see \cite{Bellet2014,Kulis2013}.

A majority of the metric learning methods
adopt Mahalanobis distance (Eq.~\eqref{eq:mahalanobis-distance}) as the distance function, 
and minimize the training loss under various constraints.
As mentioned earlier, these methods do not make distinction between
unlabeled (query) objects and labeled objects,
in the sense that their coordinates are transformed by the same matrix, $\mat{L}$ in Equation~\eqref{eq:mahalanobis-distance}.
Our approach differs from these methods in that it projects only the labeled objects
to new coordinates.

There are various strategies for learning Mahalanobis distance.
Xing et~al.~\cite{Xing2002} formulated metric learning as a convex optimization problem,
and demonstrated its effectiveness in clustering tasks.
The \emph{large-margin nearest neighbor} (LMNN) method \cite{Weinberger2009}
is probably the most popular of all metric learning methods.
Its objective is to minimize distances between objects with the same label,
and to penalize objects with different labels when they are closer than a certain distance.
Hence objects from different classes are separated by a large margin.
To make the problem convex,
in Xing et~al.'s method and LMNN, optimization is done over not $\mat{L}$ but $\mat{M}=\mat{L}^{\text{T}}\mat{L}$, with semidefinite programming.
Ying and Li \cite{Ying2012} presented an eigenvalue optimization framework for learning Mahalanobis distance.
Davis et al.~\cite{Davis2007} proposed \emph{information-theoretic metric learning} (ITML).
ITML minimizes the LogDet divergence subject to linear constraints. 
It thus requires no eigenvalue computation or semi-definite programming.

Although it has been shown that these methods work well
in many applications,
learning Mahalanobis distance typically incurs high computational cost.
Indeed, as we show in an experiment (Section~\ref{sec:experiments}), 
these methods spend substantial time in optimizing $\mat{M}$, 
when applied to large datasets.

\section{Proposed method}
\label{sec:proposed-method}

In this section, we present our approach for improving the $k$-NN classification accuracy.

In nearly all metric learning methods, the objective function to be optimized
involves a term that encourages objects of the same class to be placed closer.
In the same vein,
our method also optimizes the transformation matrix $\mat{W}$ in Eq.~\eqref{eq:backward-mapping} by
minimizing the distance between objects of the same class.
However, in our formulation, the learned transformation $\mat{W}$ is only applied to labeled objects.

Our training procedure consists of two steps.
We first make training object pairs for which the distance should be minimized.
To this end,
we follow Weinberger and Saul~\cite{Weinberger2009}:
for each labeled object $\mat{x}_i \in \mathbb{R}^d$ in the training set, we define its ``target'' objects $\mathcal{T}_i$ to be
the $k$ objects in the training set that belong to the same class as $\mat{x}_i$ and are closest to $\mat{x}_i$
as measured by the original Euclidean distance (i.e., the one before training).
We then find a matrix $\mathbf{W} \in \mathbb{R}^{d \times d}$
that moves objects in $\mathcal{T}_i$ towards $\mat{x}_i$,
by solving the following optimization problem:
\begin{equation}
  \min_\mathbf{W} \sum_{i=1}^{n} \sum_{\mat{z}\in \mathcal{T}_i} \| \mathbf{x}_i - \mathbf{W} \mathbf{z} \|^2 + \lambda \| \mathbf{W} \|^2_\mathrm{F}, \label{eq:target-regression} 
\end{equation}
where $\lambda \ge 0$ is a hyperparameter for regularization
and $\|\cdot\|_{\text{F}}$ represents the Frobenius norm.
Equation~\eqref{eq:target-regression} is a familiar objective function of ridge regression, 
and we have the closed-form solution\footnote{See appendix for derivation.}:
\begin{equation}
  \mathbf{W} = \mathbf{X} \mathbf{J} \mathbf{X}^{\text{T}} (\mathbf{XKX}^{\text{T}} + \lambda \mathbf{I})^{-1}, \label{eq:closed-form}
\end{equation}
where $\mathbf{X} = [ \mathbf{x}_1, \dots, \mathbf{x}_n ] \in \mathbb{R}^{d \times n}$,
$\mathbf{J} \in \{0,1\}^{n \times n}$ is an indicator matrix such that 
$[\mathbf{J}]_{i,j}=1$ if $\mathbf{x}_j \in \mathcal{T}_i$ and $0$ otherwise, 
and $\mathbf{K} = \diag(\mathbf{J}^{\mathrm{T}} \mathbf{e})$ is a diagonal matrix of which
the $j$th diagonal element $[\mathbf{K}]_{j,j}$ holds the number of times sample $\mat{x}_j$ appears in the $k$-nearest neighbors of other samples.
Here, $\mathbf{e}$ is a vector of all $1$'s.

In the test phase, 
we first compute the image $\mat{x}'_i = \mat{W}\mat{x}_i$ of every labeled object $\mat{x}_i$ by the learned matrix $\mathbf{W}$.
We then carry out $k$-NN classification by regarding $\mathcal{D}' = \{ (\mat{x}'_i, y_i) \}$ as the labeled objects
in place of the original ones, $\mathcal{D} = \{ (\mat{x}_i, y_i) \}$.
In the case of $1$-NN classification, for example,
this amounts to using the dissimilarity function $f$ given by Eq.~\eqref{eq:backward-mapping}
in the decision rule of Eq.~\eqref{eq:1-nn}, i.e.,
\begin{equation}
  \hat{y}
  = \mathop{\arg \min}_{y_i : (\mat{x}'_i, y_i) \in \mathcal{D}'} \| \mat{x} - \mat{x}'_i \|^2
  = \mathop{\arg \min}_{y_i : (\mat{x}_i, y_i) \in \mathcal{D}} \| \mat{x} - \mat{W} \mat{x}_i \|^2. \label{eq:1-nn-proposed}
\end{equation}

\section{Proposed method reduces hubness}
\label{sec:hubness}

In this section, we argue that the proposed method is by design less
susceptible to producing hubs \cite{Radovanovic2010} in the transformed labeled objects.
This property is desirable, as hubs have been recognized as one of the major
factors that harm the effectiveness of nearest neighbor methods.

\subsection{Hubness phenomenon}
\label{sec:hubness-phenomenon}

The hubness phenomenon \cite{Radovanovic2010} 
states that a small number of objects in the dataset, called \emph{hubs},
may occur as the nearest neighbor of many objects. 
The presence of hubs will diminish the utility of nearest-neighbor methods, 
because
the lists of nearest neighbors frequently contain the same hub objects regardless of the query.

Hubs occur in data because of an inherent bias present in Euclidean space,
called \emph{spatial centrality} \cite{Radovanovic2010};
i.e.,
objects closest to the mean of the data tend to be the nearest neighbors of many objects.
This bias is known to grow stronger with the dimensionality of the space.

The following proposition by Shigeto et al.~\cite{Shigeto2015} quantifies the degree of spatial centrality
as a function of the dimension of the space and the variance of data distribution,
when the feature values of query and data follow zero-mean Gaussian distributions with different variances.
Let $E_{\mathcal{X}}[\cdot]$ and $\Var_{\mathcal{X}}[\cdot]$ respectively
denote the expectation and variance under a distribution $\mathcal{X}$.
\begin{Proposition}
\label{prop:spatial-centrality}
\cite[Proposition~1]{Shigeto2015}
Let $\mathbf{z}$ be a $d$-dimensional random vector
sampled i.i.d. from a normal distribution with zero means and diagonal covariance matrix $s^2\mat{I}$; i.e., 
$\mathbf{z} \sim \mathcal{Z}$, where $\mathcal{Z} = \mathcal{N}(\mathbf{0}, s^2\mathbf{I})$.
  Further let
  $\sigma = \sqrt{\Var_{\mathcal{Z}}[\|\mathbf{z}\|^2]}$
  be the standard deviation of the squared norm $\|\mathbf{z}\|^2$.
  
  Consider two fixed samples $\mathbf{z}_1$ and $\mathbf{z}_2$ of random vector $\mathbf{z}$,
  such that the squared norms of $\mathbf{z}_1$ and $\mathbf{z}_2$ are
  $\gamma \sigma$ apart.
  In other words,
  \begin{equation*}
    \|\mathbf{z}_2 \|^2 - \|\mathbf{z}_1 \|^2 = \gamma \sigma.
  \end{equation*}
  Let
  $\mathbf{x}$ be a point sampled from 
  a distribution $\mathcal{X}$ with zero mean.
 
  Then,
  the expected difference between the squared distances from $\mathbf{x}$ to $\mathbf{z}_1$ and $\mathbf{z}_2$ is given by
  \begin{align}
    \Delta 
    &=
    \E_{\mathcal{X}} \left[ \| \mathbf{x} - \mathbf{z}_2 \|^2 \right]
    -
    \E_{\mathcal{X}} \left[ \| \mathbf{x} - \mathbf{z}_1 \|^2 \right] \nonumber \\
    &= 
    \gamma s^2 \sqrt{2d}.
    \label{eq:delta-definition}
  \end{align}

\end{Proposition}

The quantity
in Eq.~\eqref{eq:delta-definition} 
can be interpreted as the degree of the spatial centrality bias present in the data, which causes hub formation.
If $\mathbf{z}_1$ is closer to the origin (data mean) than $\mathbf{z}_2$ is, $\Delta > 0$ because in this case $\gamma > 0$.

This implies that a query object $\mathbf{x}$ sampled from $\mathcal{X}$ is more likely to be closer to object $\mathbf{z}_1$ than to $\mathbf{z}_2$;
i.e., given query object $\mathbf{x}$, $\mathbf{z}_1$ is more likely to become its nearest neighbor.
Because this reasoning applies to any pair of objects $\mat{z}_1$ and $\mat{z}_2$ in the dataset,
it can be concluded that the objects closest to the data mean is most likely to be a hub.

Further, the factor $s^2$ in Eq.~\eqref{eq:delta-definition} suggests that, for a fixed query distribution $\mathcal{X}$,
the data distribution $\mathcal{Y}$ with smaller variance $s$ is preferable to reduce spatial centrality,
and hence hubness as well.

\subsection{Hubness and the proposed method}
\label{sec:regression}

Ridge regression reduces the variance of mapped feature values (observables)
relative to that of target (response) variables;
see, for example, Shigeto et~al. \cite[Proposition~2]{Shigeto2015}.
Thus, in our model of Eq.~\eqref{eq:target-regression},
the variance of the components of the mapped objects $\mat{W}\mat{z}$ tends to be smaller than that of $\mat{x}$.
From the discussion on hubness in Section~\ref{sec:hubness-phenomenon},
reducing the variance of data objects (which correspond to the image $\mat{W}\mat{z}$ of the labeled objects $\mat{z}$
in the proposed method)
relative to the query (unlabeled object $\mat{x}$) can reduce the spatial centrality.
By combining these arguments, we expect that the proposed approach should alleviate the emergence of hubs,
and, consequently, improve the accuracy of $k$-NN classification.

Note that we could think of a different regression problem in which query object $\mat{x}$,
not labeled object $\mat{z}$, is mapped to new coordinates:
\begin{equation}
  \min_\mathbf{W} \sum_{i=1}^{n} \sum_{\mat{z}\in \mathcal{T}_i} \| \mathbf{W} \mathbf{x}_i -  \mathbf{z} \|^2 + \lambda \| \mathbf{W} \|^2_\mathrm{F}. \label{eq:different-regression} 
\end{equation}
This would result in function $f$ as follows:
\begin{align}
  f(\mathbf{x}, \mathbf{z}) &= \| \mathbf{W}\mathbf{x} - \mathbf{z}\|^2. \label{eq:forward-mapping} 
\end{align}
However, this dissimilarity function is useless as it actually \emph{promotes} hubness.
The variance of the transformed query objects shrinks as a result of regression.
Thus, in this model,
the variance of the labeled objects is made larger than the transformed query objects,
but this is not a desirable situation according to Proposition~\ref{prop:spatial-centrality}.
We also verify this in one of the experiments in Section~\ref{sec:experiments}.

\section{Experiments}
\label{sec:experiments}

We evaluate the proposed approach on various classification tasks. The objective of these experiments is to investigate whether the proposed approach can reduce the emergence of hubs, and improve the performance of $k$-NN classification.
The performance is measured against several popular metric learning methods.

\subsection{Experimental setups}
\label{sec:experimental-setup}

\subsubsection{Dataset description}

\begin{table*}[tb]
  \centering

  \caption{Dataset statistics.
    In document and image datasets, ``original dim.'' indicates the number of raw dimensions before applying PCA.
  }
  \label{tab:data}
  
  \subfloat[
    UCI datasets.
  ]{
    \label{tab:data-uci}
    \begin{tabular}{l *{5}{@{\hspace{2mm}}r}}
      \toprule
      dataset    & ionosphere & balance-scale & iris & wine & glass \\
      \midrule                                                                                  
      \#objects  & 351        & 625           & 150  & 178  & 214   \\
      \#classes  & 2          & 3             & 3    & 3    & 6     \\
      dimension  & 34         & 4             & 4    & 13   & 9     \\
      \bottomrule
    \end{tabular}
  }

  \subfloat[
    Document datasets.
  ]{
  \label{tab:data-document}
    \begin{tabular}{l *{4}{@{\hspace{2mm}}r}}
      \toprule
      dataset       & RCV   & News  & Reuters & TDT    \\
      \midrule                                                                                  
      \#objects     & 9625  & 18846 & 8213    & 10021  \\
      \#classes     & 4     & 20    & 41      & 56     \\
      dimension     & 300   & 300   & 300     & 300    \\
      original dim. & 29992 & 26214 & 18933   & 36771 \\
      \bottomrule
    \end{tabular}
  }
  \hspace{2mm}
  \subfloat[
    Image datasets.
  ]{
    \label{tab:data-image}
    \begin{tabular}{l *{4}{@{\hspace{2mm}}r}}
      \toprule
      dataset       & AwA   & CUB   & SUN   & aPY   \\
      \midrule                                                                                  
      \#objects     & 30475 & 11788 & 14340 & 15339 \\
      \#classes     & 50    & 200   & 717   & 32    \\
      dimension     & 300   & 300   & 300   & 300   \\
      original dim. & 4096  & 4096  & 4096  & 4096  \\
      \bottomrule
    \end{tabular}
  }

\end{table*}

Three types of datasets were used for our evaluation: UCI, document, and image datasets.

From the UCI machine learning datasets,\footnote{\url{http://archive.ics.uci.edu/ml/}}
we chose balance-scale, glass, ionosphere, iris, and wine,
as they are frequently used for evaluation in metric learning literature \cite{Ying2012,Weinberger2009,Davis2007,Jain2012}.
However, they are mostly toy problems,
and
their small feature dimensions, the numbers of labels and objects do not necessarily reflect real-world problems.
We therefore used document and image datasets also for our evaluation.

For document and image classification, support vector machines are known to provide
state-of-the-art accuracy.
Notice, however, that our goal is not to design a state-of-the-art classifier. 
Rather, the main objective of this experiments is to evaluate the performances of the proposed method in comparison with metric learning methods,
and to show its usefulness for $k$-NN classification. 

For document classification tasks,
we used four publicly available document datasets:
RCV1-v2 (RCV), 20 newsgroups (News), Reuters21578 (Reuters), and TDT2 (TDT).\footnote{Datasets were downloaded from \url{http://www.cad.zju.edu.cn/home/dengcai/Data/TextData.html}}
In Reuters21578 and TDT2, we removed minority classes that hold less than 10 objects in the dataset.
After this removal, Reuters21578 and TDT2 had 56 and 41 classes, respectively.

For image classification,
we used the following image datasets: aPascal \& aYahoo (aPY), Animals with Attributes (AwA), Caltech-UCSD Birds-200-2011 (CUB), and
SUN Attribute.\footnote{We used the publicly available features from \url{https://zimingzhang.files.wordpress.com/2014/10/cnn-features1.key}}

The computational cost of metric learning methods is heavily dependent on the dimension of the feature space.
In our preliminary experiment, 
training of the metric learning methods (LMNN, ITML, and DML-eig; see below)
did not complete in a reasonable time on document and image datasets.
We therefore had to use principal component analysis
to reduce the dimensionality of features to 300 for these datasets.\footnote{
We also conducted experiments where the dimensionality of features was set to 100. 
The results are not presented here due to lack of space, but the same trend was observed.
}

The dataset statistics are summarized in Table~\ref{tab:data}.

All data (set of feature vectors) were centered before training.
For the wine dataset, we further converted the features to z-scores,
following the remark on the UCI website that a $k$-NN classifier achieved a high accuracy with this standardization.

Each dataset was randomly split into training (70\%) and test (30\%) sets. 
Experiments were repeated on four different random splits, for which we report the average performance.

\subsubsection{Compared methods}

We trained distance/dissimilarity functions using the following methods,
and carried out $k$-NN classification on the datasets above.
\begin{itemize}
\item original metric: Euclidean distance in the original feature space, without any training.
This is the baseline.

\item LMNN: Large margin nearest neighbor classification~\cite{Weinberger2009}. 
  This method has often been used in distance metric learning experiments as a baseline.

\item ITML: Information theoretic metric learning~\cite{Davis2007}. 

\item DML-eig: Distance metric learning with eigenvalue optimization~\cite{Ying2012}.

\item \movelabeled{} (proposed method): Learning to move labeled objects toward query objects. 
  This is our proposed approach that optimizes the ridge regression objective of Eq.~\eqref{eq:target-regression},
  and predicts the labels using Eq.~\eqref{eq:1-nn-proposed}.

\item \movequery{}: Learning to move query (unlabeled) objects toward fixed labeled objects.
  This is the method discussed in Section~\ref{sec:regression} and optimizes Eq.~\eqref{eq:different-regression}.
  Like \movelabeled{} it is also based on ridge regression, but the roles of the input and response variables are exchanged.
  The resulting dissimilarity function of Eq.~\eqref{eq:forward-mapping} is then used for
  $k$-NN classification.
\end{itemize}

Notice that \movequery{} was tested only to verify the claim made in Section~\ref{sec:regression};
i.e., although both \movelabeled{} and \movequery{} are based on ridge regression,
the proposed method, \movelabeled{}, is expected to perform well by reducing hubness,
whereas \movequery{} is expected to do the contrary, by \emph{promoting} hubness.
It was therefore not meant as a competitive method.

LMNN, ITML, and DML-eig learn a Mahalanobis distance.
For these methods,
we used the publicly available MATLAB implementations provided by the respective authors\footnote{LMNN: \url{https://bitbucket.org/mlcircus/lmnn/downloads},\\
  ITML: \url{http://www.cs.utexas.edu/~pjain/itml/},\\
  DML-eig: \url{http://www.albany.edu/~yy298919/software.html}}.
We implemented the proposed method also in MATLAB\footnote{This code will be made available at our homepage.},
for fair evaluation of running time.

For LMNN, \movelabeled{}, and \movequery{}, the number of target objects for each training object was set to $1$;
i.e., for each object $\mat{x}_i$ in the training set,
we made a training pair $(\mat{x}_i, \mat{z})$ whose distance should be minimized,
where $\mat{z}$ is the object nearest to $\mat{x}_i$ among those with the same class label as $\mat{x}_i$ in the training set,
with the distance measured by the original Euclidean metric.
For the parameters of ITML on UCI datasets,
the default values in the authors' implementation were used,
and for document and image datasets, we followed Jain et~al. \cite{Jain2012}.
For DML-eig, the default setting in the authors' code was used for obtaining pairwise constraints.
We calibrated the parameter $k$ of $k$-NN classification to be used at the test time
and all other parameters ($\gamma$ in ITML, $\mu$ in DML-eig, and $\lambda$ in \movelabeled{} and \movequery)
by cross validation on the training set.

\subsubsection{Evaluation Criteria}

The methods were evaluated in three respects: 
(i) the accuracy of $k$-NN classification using the distance/dissimilarity measure learned by each method,
(ii) training time, and (iii) the degree of hubness in the data with respect to the learned distance/dissimilarity.

Following the literature \cite{Radovanovic2010,Schnitzer2012,Suzuki2013,Hara2014,Shigeto2015}, 
we used the skewness of $N_{10}$ distribution as the measure of hubness in the data.
The $N_{10}$ distribution is the empirical distribution of
the number $N_{10}(i)$ of times
each labeled object $i$ is found in the $10$-nearest neighbors of query (unlabeled) objects,
and its skewness is defined as follows:
\begin{equation*}
  \text{($N_{10}$ skewness)} = \frac{\mathbb \sum_{i=1}^{n} \left( N_{10}(i) - \E  \left[ N_{10} \right] \right)^3 / n }{ \Var \left[  N_{10} \right]^\frac{3}{2}}
\end{equation*}
where
$n$ is the total number of labeled objects,
and $\E[N_{10}]$ and $\Var[N_{10}]$ are respectively the empirical mean and variance of $N_{10}(i)$ over $n$ labeled objects.
A large $N_{10}$ skewness value indicates
the existence of labeled objects that frequently appear in the $10$-nearest neighbor lists of query objects,
i.e., hubs.

\subsection{Experimental results and discussion}
\label{sec:experimental-result}

\begin{table*}[tb]
  \centering

  \caption{Skewness of $N_{10}$ distribution: 
    A high skewness indicates the emergence of hubs (smaller is better). 
    The bold figure indicates the best performer.
  }
  \label{tab:skewness}
  
  \subfloat[
    UCI datasets.
  ]{
    \label{tab:skewness-uci}
    \begin{tabular}{l *{5}{@{\hspace{1.8em}}r}}
      \toprule
      method             & ionosphere    & balance-scale & iris          & wine            & glass \\
      \midrule                                                                                  
      original metric    & 1.65          & 0.93          & 0.40          & 0.71            & 0.77\\
      LMNN               & 1.05          & 0.63          & 0.39          & 0.61            & 0.74 \\ 
      ITML               & 0.96          & 0.79          & \textbf{0.10} & 0.43            & 0.70 \\ 
      DML-eig            & \textbf{0.78} & 0.66          & 0.41          & \textbf{0.38}   & \textbf{0.59} \\ 
      \movelabeled{} (proposed) & 1.04          & \textbf{0.56} & 0.32          & 0.55            & \textbf{0.59} \\
      \movequery{}          & 1.67          & 1.13          & 0.32          & 0.89            & 1.18 \\ 
      \bottomrule
    \end{tabular}
  }
  
  \subfloat[
    Document datasets.
  ]{
    \label{tab:skewness-document}
    \begin{tabular}{l *{4}{@{\hspace{2em}}r}}
      \toprule
      method             & RCV & News & Reuters & TDT \\
      \midrule                                                                                  
      original metric    & 13.35 & 21.93 & 7.61 & 4.89\\ 
      LMNN               & 3.86 & 14.74 & 7.63 & 4.01\\ 
      ITML               & 4.27 & 19.65 & 7.30 & 2.39\\ 
      DML-eig            & 1.71 & \textbf{1.45} & \textbf{3.05} & \textbf{1.34}\\ 
      \movelabeled{}  (proposed)  & \textbf{1.14} & 2.88 & 4.53 & 1.44\\ 
      \movequery{}          & 21.57 & 33.36 & 17.49 & 6.71\\  
      \bottomrule
    \end{tabular}
  }

  \subfloat[
    Image datasets.
  ]{
    \label{tab:skewness-image}
    \begin{tabular}{l *{4}{@{\hspace{2em}}r}}
      \toprule
      method      & AwA & CUB & SUN & aPY  \\
      \midrule                                                                                  
      original metric   & 2.49 & 2.38 & 2.52 & 2.80\\
      LMNN              & 3.10 & 2.96 & 2.80 & 3.94\\ 
      ITML              & 2.42 & 2.27 & 2.37 & 2.69\\
      DML-eig           & 1.90 & 1.77 & 2.39 & 2.17\\ 
      \movelabeled{} (proposed) & \textbf{1.24} & \textbf{0.97} & \textbf{1.02} & \textbf{1.23}\\
      \movequery{}         & 7.81 & 7.83 & 7.48 & 11.65\\
      \bottomrule
    \end{tabular}
  }
  
\end{table*}

\begin{table*}[tb]
  \centering

  \caption{Classification accuracy [\%]:
    Bold figures indicate the best performers for each dataset.
  }
  \label{tab:accuracy}
  
  \subfloat[
    UCI datasets.
  ]{
    \label{tab:accuracy-uci}
    \begin{tabular}{l *{5}{@{\hspace{1.8em}}r}}
      \toprule
      method                           & ionosphere    & balance-scale & iris          & wine          & glass         \\
      \midrule
      original metric & 86.8          & 89.5          & 97.2          & 98.1          & 68.1          \\
      LMNN            & \textbf{90.3} & 90.0          & 96.7          & 98.1          & 67.7          \\
      ITML            & 87.7          & 89.5          & \textbf{97.8} & \textbf{99.1} & 65.0          \\ 
      DML-eig         & 87.7          & \textbf{91.2} & 96.7          & 98.6          & 66.5          \\
      \movelabeled{} (proposed) & 89.6          & 89.5          & 97.2          & 98.6          & \textbf{70.8} \\
      \movequery{}       & 79.7          & 89.4          & 97.2          & 96.3          & 62.3          \\
      \bottomrule
    \end{tabular}
  }
  
  \subfloat[
    Document datasets.
  ]{
    \label{tab:accuracy-document}
    \begin{tabular}{l *{4}{@{\hspace{2em}}r}}
      \toprule
                                         method          & RCV           & News          & Reuters       & TDT          \\
      \midrule
                                         original metric & 92.1          & 76.9          & 89.5          & 96.1          \\
                                         LMNN            & \textbf{94.7} & 79.9          & 91.5          & 96.6          \\
                                         ITML            & 93.2          & 77.0          & 90.8          & 96.5          \\ 
                                         DML-eig         & 94.5          & 73.3          & 85.9          & 95.7          \\
                                         \movelabeled{} (proposed) & 94.4          & \textbf{81.6} & \textbf{91.6} & \textbf{96.7}   \\
                                         \movequery{}        & 89.1          & 70.0          & 85.9          & 95.4          \\
      \bottomrule
    \end{tabular}
  }

  \subfloat[
    Image datasets.
  ]{
    \label{tab:accuracy-image}
    \begin{tabular}{l *{4}{@{\hspace{2em}}r}}
      \toprule
                                         method          & AwA           & CUB           & SUN           & aPY           \\
      \midrule
                                         original metric & 83.2          & 51.6          & 26.2          & 82.2          \\
                                         LMNN            & 83.0          & \textbf{54.7} & 24.4          & 81.8          \\
                                         ITML            & 83.1          & 51.3          & 26.0          & 82.4          \\  
                                         DML-eig         & 82.0          & 53.5          & 22.4          & 81.6          \\   
                                         \movelabeled{} (proposed) & \textbf{84.1} & 52.4          & \textbf{28.3} & \textbf{83.4}   \\
                                         \movequery{}       & 79.2          & 43.3          & 14.6          & 78.7          \\ 
      \bottomrule
    \end{tabular}
  }
  
\end{table*}

\subsubsection{Skewness}

Table~\ref{tab:skewness} shows the skewness of $N_{10}$ distribution.
For all datasets, 
we observe that the proposed approach (\movelabeled{}) reduced $N_{10}$ skewness considerably compared with the original Euclidean distance,
meaning that it effectively suppressed the emergence of hub objects.
$N_{10}$ skewness was reduced by metric learning methods (LMNN, ITML, and DML-eig) on many datasets, most notably by DML-eig. 
Also, as expected from the discussion of Section~\ref{sec:regression},
\movequery{} increased $N_{10}$ skewness except for the iris dataset.

\subsubsection{Accuracy}

Tables~\ref{tab:accuracy} shows the classification accuracy.
In most datasets, both the metric learning methods and the proposed method
outperformed the original distance metric.
The proposed method is comparable with, or slightly better than, the metric learning methods. 
Although \movequery{} optimized the minimizing distance between objects in same class (our proposed method also optimized such distance),
the method obtained poor results even compared with the original Euclidean metric except for the iris datasets.

Note that, in UCI datasets, we observed that the proposed method did not work well,
and even \movequery{} were competitive with others.
This is an expected result, because the UCI datasets did not have much hubness even with the original metric
(see Table~\ref{tab:skewness-uci}).
Hubs tend to emerge in high dimensional space \cite{Radovanovic2010,Schnitzer2012,Hara2014},
but all the UCI datasets have a small dimensionality (see Table~\ref{tab:data-uci}).
Consequently, hub reduction/promotion methods did not affect the result significantly.

\subsubsection{Training time}

\begin{table*}[tb]
  \centering

  \caption{Training time [sec]:
    Bold figures indicate the best performer for each dataset.
  }
  \label{tab:training-time}
  
  \subfloat[
    Document datasets.
  ]{
    \label{tab:training-time-document}
    \begin{tabular}{l *{4}{@{\hspace{2.2mm}}r}}
      \toprule
                                                   method          & RCV           & News          & Reuters       & TDT                          \\
      \midrule                                                                                  
                                                   LMNN            & 1713.0        & 1164.7        & 676.2         & 886.1                         \\ 
                                                   ITML            & 35.5          & 1512.5        & 124.1         & 169.0                         \\ 
                                                   DML-eig         & 762.2         & 6145.9        & 2710.4        & 2350.6                        \\ 
                                                   proposed        & \textbf{6.0}  & \textbf{7.0}  & \textbf{4.6}  & \textbf{16.1}                 \\ 
      \bottomrule
    \end{tabular}
  }
  \hspace{2mm}
  \subfloat[
    Image datasets.
  ]{
    \label{tab:training-time-image}
    \begin{tabular}{l *{4}{@{\hspace{2.2mm}}r}}
      \toprule
                                       method          & AwA           & CUB           & SUN           & aPY \\
      \midrule                                                                                  
                                       LMNN            & 1525.5        & 1098.2        & 15704.3       & 317.3        \\ 
                                       ITML            & 1536.3        & 577.6         & 1126.4        & 9211.2       \\
                                       DML-eig         & 2048.0        & 2084.7        & 2006.1        & 1787.1       \\
                                       proposed        & \textbf{9.5}  & \textbf{1.5}  & \textbf{4.1}  & \textbf{6.4} \\
      \bottomrule
    \end{tabular}
  }
  
\end{table*}

To investigate the computational cost,
we measured the elapsed real time needed to train the proposed method and the metric learning methods.

Table~\ref{tab:training-time} shows the average training time in document and image datasets.
We observe that the proposed approach has a clear advantage in terms of training cost.
It was faster than any metric learning methods compared.
Indeed, on all datasets except RCV, 
it was more than two orders of magnitude faster than the fastest metric learning methods.
This can be explained by the fact that
the metric learning methods take burden of optimizing over Mahalanobis metric.
To enforce the constraint that the matrix $\mat{M}$ in Eq.~(\ref{eq:mahalanobis-distance-orig})
should remain positive semi-definite,
these methods pay high computational cost, e.g., to check the non-negativity of eigenvalues,
at every training iteration.
In contrast, the proposed approach has a closed-form solution;
although this solution depends on matrix inverse,
it needs to be calculated only once.

\section{Conclusion}
\label{sec:conclusion}

In this paper, we have proposed a simple regression-based technique to improve $k$-NN classification accuracy.
The results of our work can be summarized as follows:
\begin{itemize}
\item
  To improve $k$-NN classification accuracy,
  we proposed learning a transformation of labeled objects,
  without altering the coordinates of query (unlabeled) objects.
  This approach is justified from the perspective of reducing hubness in the labeled objects.
  Because our method is inherently designed to suppress hubness,
  it need not consider pairs of objects from different classes during training.
  The number of such pairs can be enormous and their use also renders the optimization problem non-convex,
  which is therefore a major obstacle to the scalability of metric learning methods.

\item 
  Our method deviates from the metric learning framework as the learned transformation $\mat{W}$ does not provide
  a proper metric.
  In $k$-NN classification, however,
  labeled objects can be interpreted not as mere points but rather a representation of 
  the (non-linear) decision boundaries between classes.
  Our approach follows this interpretation
  and changes the decision boundaries, through $\mat{W}$, to improve classification accuracy.

\item
  In the experiments of Section~\ref{sec:experiments},
  our approach indeed improved the classification accuracy when the data was high-dimensional and hubs emerged.
  It outperformed metric learning methods
  on most document and image datasets, and comparable on the rest.
  However, its effect was not evident on the UCI datasets,
  in which hubness was not noticeable because of the low dimensionality of the data.
  
\item
  The experiments showed that our approach was substantially faster than the compared metric learning methods.
  For large document and image datasets,
  the speed-up was more than two orders of magnitude over the fastest metric learning methods,
  although the classification accuracy was better or comparable.
\end{itemize}

We have focused on multi-class classification problems in this paper,
but hubness is known to be harmful in other situations,
such as clustering and semi-supervised classification in high-dimensional space \cite{Radovanovic2010}.
We plan to extend our approach to deal with these situations.
We will also extend our method to learn nonlinear metrics.

Another direction of future work is to investigate the effect of our approach on kernel machines.
Metric learning has been shown to be
an effective preprocessing for kernel machines \cite{Xu2013,Weinberger2007,Dhillon2010},
and
we will pursue a similar line using our approach.

\subsubsection*{Acknowledgments}

We thank anonymous reviewers of PAKDD for their comments and suggestions.
We also thank Pierre Leleux and Marco Saerens for pointing out an error
in the closed-form formula (Eq.~\eqref{eq:closed-form}) in the earlier version of the manuscript.
This work was partially supported by JSPS Kakenhi Grant no.~15H02749.

\bibliographystyle{splncs03}
\bibliography{dml,hub}

\section*{Appendix: The derivation of the closed form solution}
\label{sec:derivation-closed-form}

Let $\mathrm{Tr}(\cdot)$ denote matrix trace.
Let $\mathbf{e}$ be an $n$-dimensional vector of all $1$'s, and $\mathbf{e}_i$ be an $n$-dimensional vector with $1$ at the $i$th component and $0$ everywhere else.
Consider a set of $n$ objects, with $\mathbf{x}_i \in \mathbb{R}^d$ denoting the vector representation of object $i$ ($i = 1, \ldots, n$).
Let $\mathcal{N}$ be a set of source-target object indices, such that
$(i,j) \in \mathcal{N}$ indicates that object $j$ is one of the desired target objects for source object $i$, e.g.,
$j$ is one of the same-class objects nearest to $i$.
Let $\mathcal{N}_i = \{ j \mid (i, j) \in \mathcal{N} \}$ be the indices of the target objects for source $i$.
For brevity, we assume for every source object $i$, there are exactly $k$ target objects in $\mathcal{N}$; thus, $|\mathcal{N}_i| = k$ for $i=1, \ldots, n$.

The objective function in Eq.~\eqref{eq:target-regression} can be written as
\begin{align}
  L 
  &= \sum_{i=1}^{n} \sum_{j \in \mathcal{N}_i} \| \mathbf{x}_i - \mathbf{W} \mathbf{x}_j \|^2 + \lambda \| \mathbf{W} \|^2_\mathrm{F} \nonumber \\
  &= \sum_{i=1}^n \mathrm{Tr}[  (\mathbf{X} \mathbf{S}_i - \mathbf{W} \mathbf{X} \mathbf{T}_i )^\mathrm{T} (\mathbf{X} \mathbf{S}_i - \mathbf{W} \mathbf{X} \mathbf{T}_i ) ]+ \lambda \mathrm{Tr}[ \mathbf{W}^\mathrm{T} \mathbf{W} ] . \label{eq:target-regression-alt}
\end{align}
Here,
$\mathbf{S}_i = [ \mathbf{e}_i \, \mathbf{e}_i \cdots \mathbf{e}_i ] \in \{0,1\}^{n\times k}$ is a matrix with all $1$'s in row $i$ and $0$ everywhere else,
and
$\mathbf{T}_i = [ \mathbf{e}_{j_1} \, \mathbf{e}_{j_2} \cdots \mathbf{e}_{j_k} ]\in \{0,1\}^{n \times k}$ where $\mathcal{N}_i = \{ j_1, \ldots, j_k \}$.

Let $\mathbf{S}, \mathbf{T} \in \{0,1\}^{n \times nk}$ be such that 
$\mathbf{S} = [ \mathbf{S}_1, \dots , \mathbf{S}_n]$ and $\mathbf{T} = [ \mathbf{T}_1, \dots , \mathbf{T}_n]$.
Eq.~\eqref{eq:target-regression-alt} can further be rewritten as 
\begin{align*}
  L 
  &= \mathrm{Tr}[  (\mathbf{X} \mathbf{S} - \mathbf{W} \mathbf{X} \mathbf{T} )^\mathrm{T} (\mathbf{X} \mathbf{S} - \mathbf{W} \mathbf{X} \mathbf{T} ) + \lambda \mathbf{W}^\mathrm{T} \mathbf{W} ] \nonumber \\
  &= \mathrm{Tr}[  \mathbf{S}^\mathrm{T}\mathbf{X}^\mathrm{T} \mathbf{X} \mathbf{S} - 2 \mathbf{S}^\mathrm{T}\mathbf{X}^\mathrm{T} \mathbf{W}\mathbf{X}\mathbf{T} + \mathbf{T}^\mathrm{T}  \mathbf{X}^\mathrm{T} \mathbf{W}^\mathrm{T} \mathbf{W} \mathbf{X} \mathbf{T}
   + \lambda \mathbf{W}^\mathrm{T} \mathbf{W} ] .
\end{align*}
Taking the derivative of this function with respect to $\mathbf{W}$, we have:
\begin{align*} 
    \frac{\partial L}{\partial \mathbf{W}} 
    &= -2 \mathbf{XST}^\mathrm{T} \mathbf{X}^\mathrm{T} + 2 \mathbf{WXT} \mathbf{T}^\mathrm{T} \mathbf{X}^\mathrm{T} + 2 \lambda \mathbf{W} .
\end{align*}
Setting this to zero, 
we obtain the closed form solution given by Eq.~\eqref{eq:closed-form}, restated here as follows: 
\begin{align*}
    \mathbf{W}
    &= \mathbf{X} 	\underbrace{\mathbf{ST}^\mathrm{T}}_{\mathbf{J}} \mathbf{X}^\mathrm{T} (\mathbf{X} \underbrace{\mathbf{TT}^\mathrm{T}}_{\mathbf{K}}  \mathbf{X}^\mathrm{T} + \lambda \mathbf{I})^{-1} \nonumber \\
    &= \mathbf{X} \mathbf{J} \mathbf{X}^\mathrm{T} (\mathbf{X} \mathbf{K} \mathbf{X}^\mathrm{T} + \lambda \mathbf{I})^{-1} ,
\end{align*}
where 
$\mathbf{J} = \mathbf{S} \mathbf{T}^{\mathrm{T}}$ and 
$\mathbf{K} = \mathbf{T} \mathbf{T}^{\mathrm{T}}$. 
Note that matrix $\mathbf{J} \in \{0,1\}^{n\times n}$ is the indicator matrix of $\mathcal{N}$, such that the $(i,j)$-element is $1$ if $(i,j) \in \mathcal{N}$, and $0$ otherwise.
The other matrix $\mathbf{K}$ is a diagonal matrix where the $j$th diagonal entry represents the number of times object $j$ occurs as a target in $\mathcal{N}$, i.e., 
$[\mathbf{K}]_{j,j}  = | \{ (i,j) \in \mathcal{N} \} | = | \{ i \mid j \in \mathcal{N}_i \} | = \diag(\mathbf{J}^{\text{T}}\mathbf{e})$.

\end{document}